\documentclass[runningheads]{llncs}
\usepackage[T1]{fontenc}
\usepackage{graphicx}
\usepackage{multirow}
\usepackage{amssymb}
\usepackage{mathtools}

\begin{document}
\title{Video-based Person Re-identification with Long Short-Term Representation Learning}

\author{{Xuehu Liu}\inst{1,2} \and
	{Pingping Zhang}\inst{3} \and
	{Huchuan Lu}\inst{1,2,}\thanks{Corresponding Authors}}
\authorrunning{Liu et al.}
%
\institute{School of Information and Communication Engineering, Dalian University of Technology \and
	Ningbo Institute, Dalian University of Technology  \and
	School of Artificial Intelligence, Dalian University of Technology
	\email{snowtiger@mail.dlut.edu.cn, \{zhpp, lhchuan\}@dlut.edu.cn}}

%

%
%
%
\maketitle              
\begin{abstract}
Video-based person Re-Identification (V-ReID) aims to retrieve specific persons from raw videos captured by non-overlapped cameras.
As a fundamental task, it spreads many multimedia and computer vision applications.
However, due to the variations of persons and scenes, there are still many obstacles that must be overcome for high performance.
In this work, we notice that both the long-term and short-term information of persons are important for robust video representations.
Thus, we propose a novel deep learning framework named \textbf{Long Short-Term Representation Learning (LSTRL)} for effective V-ReID.
More specifically, to extract long-term representations, we propose a \textbf{Multi-granularity Appearance Extractor (MAE)}, in which four granularity appearances are effectively captured across multiple frames.
Meanwhile, to extract short-term representations, we propose a \textbf{Bi-direction Motion Estimator (BME)}, in which reciprocal motion information is efficiently extracted from consecutive frames.
The MAE and BME are plug-and-play and can be easily inserted into existing networks for efficient feature learning.
As a result, they significantly improve the feature representation ability for V-ReID.
Extensive experiments on three widely used benchmarks show that our proposed approach can deliver better performances than most state-of-the-arts.
%
\end{abstract}
\section{Introduction}
Video-based person Re-Identification (V-ReID) aims to retrieve specific persons from raw videos captured by non-overlapped cameras.
Due to the wide range of applications such as object tracking and scenario surveillance, V-ReID has attracted more and more attention from researchers.
During the past decade, deep learning has contributed to significant improvements in V-ReID by extracting robust and discriminative features.
Even so, the challenges of spatial occlusions and temporal misalignments in videos have yet to be solved.
To address these challenges, several efforts have been made by various representation learning methods.
For example, some works~\cite{li2018diversity,subramaniam2019co,liu2021spatial} highlight the salient visual regions of each frame, while other works~\cite{liu2019spatial,dai2019video,li2019global,liu2021watching} attentively aggregate the temporal diverse clues across different frames.
Although effective, there are still some obvious issues.
Firstly, the long-range inter-frame associations are ignored in the extraction of appearances.
Secondly, the motion information is missed out, which is instrumental in identifying persons when they have similar appearances.
Therefore, these methods struggle to extract long-term appearance representations effectively and fail to extract motion information efficiently.

\begin{figure*}[t]
	\centering
	\resizebox{0.8\textwidth}{!}
	{
		\begin{tabular}{@{}c@{}c@{}}
			\includegraphics[width=0.3\linewidth,height=0.22\linewidth]{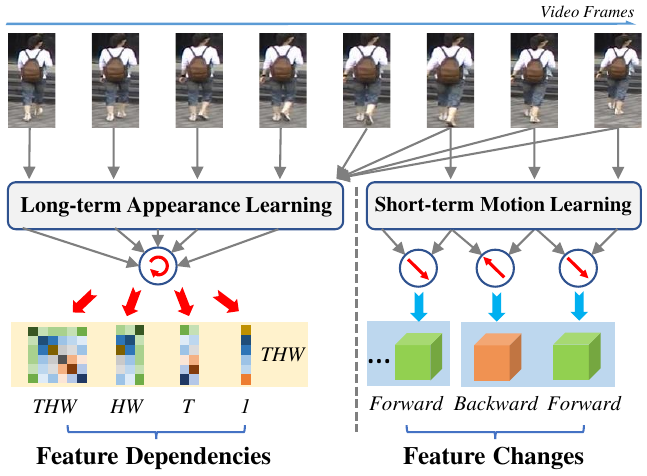} \\
		\end{tabular}
	}
	\vspace{-4mm}
	\caption{Illustration of our proposed paradigm.
		For long-term appearance learning (left part), four granularities are captured from all the frames.
		For short-term motion learning (right part), feature changes are estimated in forward and backward directions.
		The red straight arrows indicate the execution of bidirectional motion learning, while red curved arrow indicates the mining of inter-feature dependencies.
	}
	\label{fig:Intro}
	\vspace{-4mm}
\end{figure*}

In this work, we propose a novel \textbf{Long Short-Term Representation Learning (LSTRL)} framework for V-ReID.
The illustration of our LSTRL is presented in Fig.~\ref{fig:Intro}.
%
%
More specifically, to extract long-term representations, we design a \textbf{Multi-granularity Appearance Extractor (MAE)}, in which two local-to-local and two global-to-local dependencies are captured.
With these dependencies, we highlight the meaningful local features and extract multi-granularity appearances.
Further, our MAE incorporates features at multiple granularities to obtain long-term appearance representations, which contain more abundant information than the appearance of a single frame.
Meanwhile, to extract short-term representations, we design a \textbf{Bi-direction Motion Estimator (BME)}, in which the motion information is extracted from consecutive frames.
Distinct from the high-cost optical flow estimation, we reciprocally calculate the feature changes in a global-to-local manner, which brings two benefits.
The first is increasing the robustness for noise frames. The second is reducing the spatial misalignments caused by occlusion and scale variations.

Moreover, our MAE and BME are plug-and-play and can be easily inserted into existing networks for multi-stage long short-term feature learning.
To verify the effectiveness of our method, extensive experiments are conducted on three widely used V-ReID datasets.
Experimental results show that our approach performs better than most state-of-the-art methods.
In summary, the main contributions of our work are as follows:
\begin{itemize}
	\item
	We propose a novel Long Short-Term Representation Learning (LSTRL) framework for V-ReID.
	\item
	We design a Multi-granularity Appearance Extractor (MAE) to obtain long-term appearance representations, in which four kinds of dependencies are effectively explored from
	all the frames.
	\item
	We design a Bi-direction Motion Estimator (BME) to obtain short-term motion representations, in which reciprocal feature changes between consecutive frames is efficiently estimated in a global-to-local manner.
	\item
	Experimental results on three widely-used benchmarks demonstrate that our framework attains a better performance than most state-of-the-art methods.
\end{itemize}
\section{Related Work}
\subsection{Appearance Representation Learning for V-ReID}
For effective V-ReID, some works~\cite{sarfraz2018pose,hou2019vrstc,fu2019sta} aim at extracting discriminative appearance representations from spatial appearance modeling.
For example, Li \emph{et al.}~\cite{li2018diversity} mine diverse part features which are constrained by an attention regularization.
Subramaniam \emph{et al.}~\cite{subramaniam2019co} insert the co-segmentation attention module into the CNN backbone to activate salient regions of pedestrians.
Zhang \emph{et al.}~\cite{zhang2020multi} utilize the global feature to guide the extraction of multi-granularity features in each frame.
Liu \emph{et al.}~\cite{liu2021spatial} utilize human skeleton information to locate human key bodies and extract appearance information.
Hou \emph{et al.}~\cite{hou2021bicnet} a bilateral complementary network for spatial complementarity modeling.
Bai \emph{et al.}~\cite{bai2022salient} design a salient-to-broad module to enlarge the attention regions frame by frame.
These methods ignore the inter-frame associations and simply focus on extracting the appearance representations of single frame.
Different from them, we capture the long-range dependencies across all the frames and extract long-term appearance representations at multiple granularities.
\subsection{Motion Representation Learning for V-ReID}
Apart from appearance representations, the motion representation is another important clue for identifying persons.
Recently, several efforts are explored to extracting motion features from videos.
For example, McLaughlin \emph{et al.}~\cite{mclaughlin2016recurrent} extract optical flows to represent motion information.
Liu \emph{et al.}~\cite{liu2017video} accumulate motion clues of persons by recurrent feature aggregations.   
Li \emph{et al.}~\cite{li2019global} learn the global-local temporal representation to exploit the multi-scale temporal cues from video sequences.
Li \emph{et al.}~\cite{li2020appearance} utilize a generative model to predict the walking patterns of persons.
Chen \emph{et al.}~\cite{chen2020temporal} extract the coherence and motion features by temporal disentangling.
%
Liu \emph{et al.}~\cite{liu2021watching} design a temporal reciprocal learning mechanism to model disentangled video cues.
Besides, Gu \emph{et al.}~\cite{gu2022motion} model motion information based on the position and appearance changes of bodies.
Different from their high-cost motion estimation, we perform a global-to-local and reciprocal motion estimator, which significantly reduces the computational complexity and increases the robustness to noise frames and spatial misalignment.
\section{Our Method}
\begin{figure*}[t]
	\centering
	\resizebox{0.9\textwidth}{!}
	{
		\begin{tabular}{@{}c@{}c@{}}
			\includegraphics[width=\linewidth,height=0.3\linewidth]{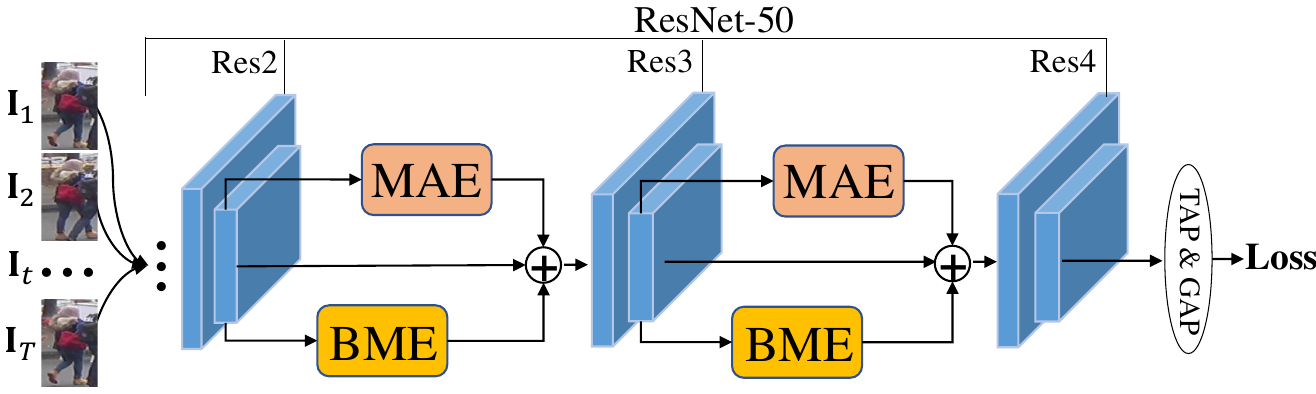} \\
		\end{tabular}
	}
	\vspace{-4mm}
	\caption{The overall framework of our LSTRL.
	}
	\label{fig:Framework}
	\vspace{-4mm}
\end{figure*}

Our proposed framework is presented in Fig.~\ref{fig:Framework}.
Given a video of persons, the Restricted Random Sampling (RRS)~\cite{li2018diversity} is first adopted to generate $T$ frames $\{\textbf{I}_1, \textbf{I}_2, ..., \textbf{I}_T\}$.
Then, we utilize a 2D CNN, such as ResNet-50~\cite{he2016deep}, to encode the multi-scale local information of each frame.
After that, our Multi-granularity Appearance Extractor (MAE) associates all the frames to extract long-term appearance representations.
Meanwhile, our Bi-direction Motion Estimator (BME) links consecutive frames to extract short-term motion representations.
The obtained representations from MAE and BME are added to frame-level features for subsequent feature extraction.
Finally, the Global Average Pooling (GAP) and Temporal Average Pooling (TAP) are deployed after the last layer to obtain a video-level representation for retrieval.
For the model training, we combine a cross-entropy loss and a batch-hard triplet loss~\cite{hermans2017defense} for end-to-end supervision.
The details of MAE and BME are described in the following sections.

\subsection{Multi-granularity Appearance Extractor}
\begin{figure}[t!]
	\centering
	\resizebox{0.9\textwidth}{!}
	{
		\begin{tabular}{@{}c@{}c@{}}
			\includegraphics[width=\linewidth,height=0.7\linewidth]{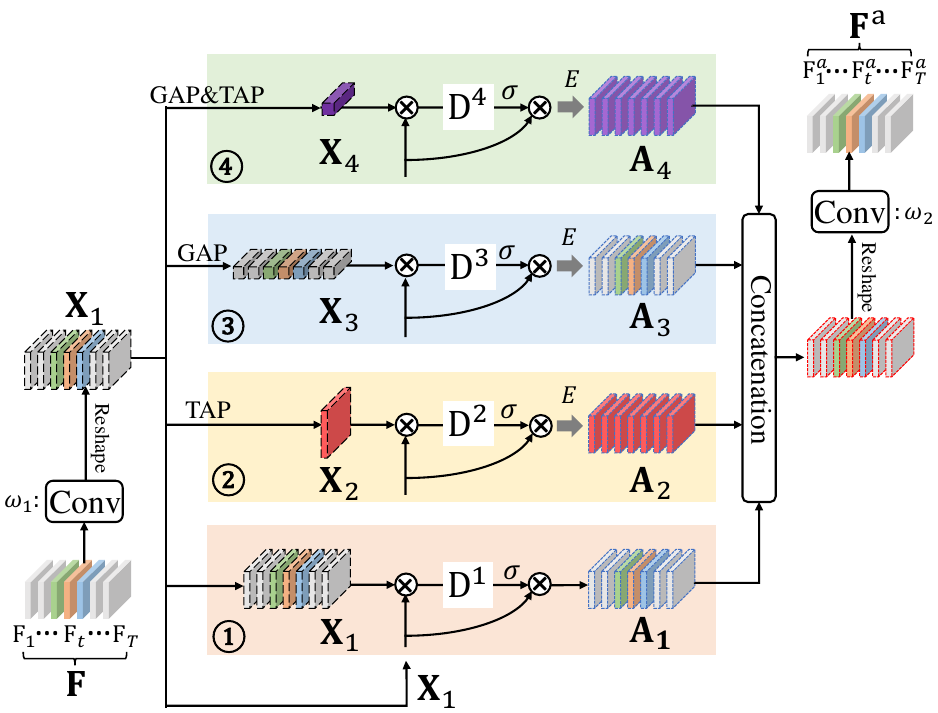} \\
		\end{tabular}
	}
	\vspace{-4mm}
	\caption{The multi-granularity appearance extractor.
	}
	\label{fig:Ablock}
	\vspace{-4mm}
\end{figure}

The detailed structure of our MAE is shown in Fig.~\ref{fig:Ablock}.
In MAE, four kinds of dependencies are captured respectively, including two local-local dependencies and two global-local dependencies.
They are beneficial to associate meaningful local features across all the frames.
With these dependencies, we can effectively extract multiple appearance features in multi-granularity, which are further incorporated to obtain long-term representations.

More specifically, after the $l$-th residual block of ResNet-50, we obtain the frame-level feature maps $\{\textbf{F}_1, \textbf{F}_2, ..., \textbf{F}_T \}$.
These frame-level features are concatenated to get $\textbf{F} \in \mathbb{R}^{T\times H\times W\times C }$ as the input of MAE, where $H$, $W$, and $C$ represent the height, weight and channels, respectively.
Then, $\textbf{F}$ is passed into one $1\times 1$ convolutional layer (${\omega}_1$) for reducing its channels to one quarter.
We reshape the channel-reduced feature and get one granularity feature as $\textbf{X}_1 \in \mathbb{R}^{THW\times \frac{1}{4}C}$.
Meanwhile, the GAP or TAP is deployed on $\textbf{X}_1$ to generate multi-granularity features $\textbf{X}_i$ ($i=2, 3, 4$).
Afterwards, as shown in Fig.~\ref{fig:Ablock}, four kinds of dependencies are generated by
\begin{equation}\label{1}
	\textbf{D}^{i} = \sigma ( \textbf{X}_i \textbf{X}^{\top}_{1} ), i=1,2,3,4,
\end{equation}
where $\top$ denotes the transpose operator, $\sigma$ is the softmax activation function.
It is noted that, the dependencies $\textbf{D}^1 \in \mathbb{R}^{THW\times THW}$ and $\textbf{D}^2 \in \mathbb{R}^{HW\times THW}$ contain the local-to-local associations, which help to enhance local features. 
The dependencies $\textbf{D}^3 \in \mathbb{R}^{T\times THW}$ and $\textbf{D}^4 \in \mathbb{R}^{1\times THW}$ contain the global-to-local associations, which help to highlight discriminative local features under global guidance.
With these dependencies, we obtain multi-granularity appearance representations by
\begin{align}\label{1}
	\textbf{A}_i =
	\begin{cases}
		\textbf{D}^{i} \textbf{X}_1,  & i=1\\
		E (\textbf{D}^{i} \textbf{X}_1 ),  & i=2,3,4\\
	\end{cases}
\end{align}
where $E$ means the feature extension operation.
For example, $\textbf{D}^{i} \textbf{X}_1 \in \mathbb{R}^{HW\times \frac{1}{4}C}$ is first reshaped as $\mathbb{R}^{1\times HW\times \frac{1}{4}C}$ and then replicated $T$ times to be $\mathbb{R}^{T\times HW\times \frac{1}{4}C}$.
After that, we concatenate its temporal dimension to obtain $\textbf{A}_2 \in \mathbb{R}^{THW\times \frac{1}{4}C}$.

%
Finally, the multi-granularity appearance representations are concatenated and weighted by
\begin{equation}\label{2}
\textbf{F}^{a} = \omega_{2} ([\textbf{A}_{1}, \textbf{A}_{2}, \textbf{A}_{3}, \textbf{A}_{4}]).
\end{equation}
where $[,]$ means the concatenation along channels, and $\omega_2$ is another $1\times 1$ convolutional layer.
In this way, we obtain the long-term appearance representation $\textbf{F}^{a} \in \mathbb{R}^{T\times H \times W \times C} $.
Compared with single-frame learning, our MAE can effectively associate meaningful local features from all the frames.
In contrast to 3D CNN-based methods, our MAE is more efficient to extract long-term representations with multi-granularity dependencies.

\subsection{Bi-direction Motion Estimator}
\begin{figure}[t]
	\centering
	\resizebox{0.8\textwidth}{!}
	{
		\begin{tabular}{@{}c@{}c@{}}
			\includegraphics[width=\linewidth,height=0.55\linewidth]{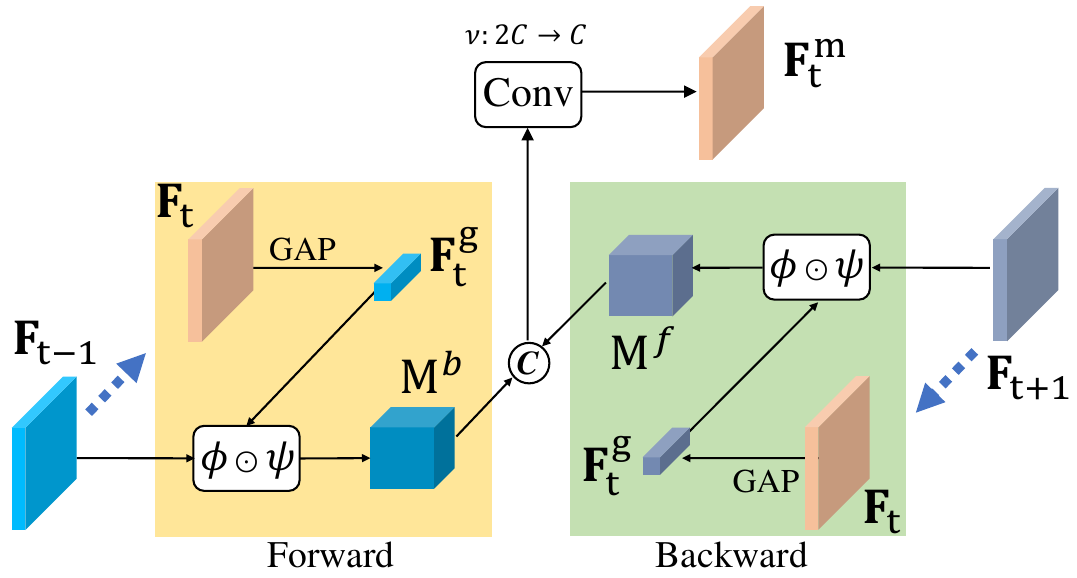} \\
		\end{tabular}
	}
	\vspace{-2mm}
	\caption{The bi-direction motion estimator.
	}
	\label{fig:Mblock}
	\vspace{-4mm}
\end{figure}

In addition, we propose a BME to extract short-term motion representations as a supplement to the appearance representations.
The structure of our BME is shown in Fig.~\ref{fig:Mblock}. 
After the $l$-th residual block of ResNet-50, our BME estimates feature changes between consecutive frames $\textbf{F}_{t-1}$, $\textbf{F}_{t}$ and $\textbf{F}_{t+1}$.

Generally, the optical flow estimation is often used for motion prediction~\cite{ilg2017flownet,sun2018pwc,zhao2020maskflownet}.
However, the pixel-to-pixel motion estimation is high-cost and suffers from severe spatial misalignment.
Different from the pixel-to-pixel motion estimation, in our BME, we calculate motion features in a global-to-local manner, which will significantly reduces the computational complexity.
Besides, the reciprocal estimation of motion information could increase the robustness to noise frames.
%
More specifically, the global feature $\textbf{F}_{t}^g \in \mathbb{R}^{1\times 1 \times C}$ is directly generated from $\textbf{F}_{t} \in \mathbb{R}^{H\times W \times C}$ by a GAP.
Then, instead of the pixel-to-pixel estimation, we estimate the motion information in a global-to-local manner, 
\begin{equation}\label{5}
	\textbf{M}^f =  \phi ( \textbf{F}_{t}^g ) \odot \psi( \textbf{F}_{t+1} ),
\end{equation}
\begin{equation}\label{5}
	\textbf{M}^b =  \phi ( \textbf{F}_{t}^g ) \odot \psi( \textbf{F}_{t-1} ),
\end{equation}
where $\odot$ presents the dot product.
$\phi$ and $\psi$ are two $1\times 1$ convolutional layers, reducing the channels to half.
After that, we concatenate $\textbf{M}^f$ and $\textbf{M}^b$, and aggregate the bi-direction motion information by
\begin{equation}\label{5}
	\textbf{F}^{m}_t = \upsilon ([\textbf{M}^f, \textbf{M}^b]),
\end{equation}
where $\upsilon$ means one $1\times 1$ convolutional layer with ReLU.
The bi-direction aggregation can realize better temporal learning, resulting in more robust motion features.
By our BME, the short-term motion representation $\textbf{F}^{m}_t$ can be efficiently extracted with reciprocal learning.

\begin{table}[t]
	\centering
	\caption{Performance (\%) comparison with state-of-the-arts on MARS and iLIDS-VID datasets. Compared methods are separated into three groups, \emph{i.e.}, spatial learning (S), temporal learning (T), spatial-temporal (ST) learning. The texts in bold and underline highlight the best and second performances.}
	\label{tab:Sota1}
	\resizebox{0.6\textwidth}{!}{%
		\begin{tabular}{c|l|cc|cc}
			\hline
			\multirow{2}{*}{}  & \multirow{2}{*}{Methods}                   & \multicolumn{2}{c|}{MARS}     & \multicolumn{2}{c}{iLIDS-VID} \\
			&                                    & mAP              & R-1              & R-1              & R-5           \\ \hline
			\multirow{6}{*}{S}  & DRSA~\cite{li2018diversity}        & 65.8             & 82.3             & 80.2             & --            \\
			& COSAM~\cite{subramaniam2019co}     & 79.9             & 84.9             & 77.8             & 97.3          \\
			& RAFA~\cite{zhang2020multi}         & 85.9             & 88.8             & 88.6             & 98.0          \\
			& PSTA~\cite{wang2021pyramid}        & 85.8             & \underline{91.5} & 91.5             & 98.1          \\
			& BiCNet~\cite{hou2021bicnet}        & 86.0             & 90.2             & --               & --            \\
			& SINet~\cite{bai2022salient}        & 86.2             & 91.0             & \textbf{92.5} & --            \\ \hline
			\multirow{7}{*}{T}  & RNN+Flow ~\cite{mclaughlin2016recurrent} & --            & --            & 58      & 84                  \\
			& AMOC+Flow~\cite{liu2017video}    & --               & --               & 68.7             & 94.3          \\
			& AMEM~\cite{li2020appearance}       & 79.3             & 86.7             & 87.2             & 97.7          \\
			& GLTR~\cite{li2019global}           & 78.5             & 87.0             & 86.0             & --            \\
			& TCLNet~\cite{hou2020temporal}      & 85.1             & 89.8             & 86.6             & --            \\
			& GRL~\cite{liu2021watching}         & 84.8             & 91.0             & 90.4             & \underline{98.3}          \\ \hline
			\multirow{6}{*}{ST} & M3D~\cite{li2019multi}             & 74.1             & 84.4             & 74.0             & 94.3          \\
			& AP3D~\cite{gu2020appearance}       & 85.6             & 90.7             & 88.7             & --            \\
			& SSN3D~\cite{jiang2021ssn3d}        & 86.2             & 90.1             & --               & --            \\
			& STMN~\cite{eom2021video}           & 84.5             & 90.5             & --               & --            \\
			& CTL~\cite{liu2021spatial}          & \underline{86.7} & 91.4             & 89.7             & 97.0          \\
			& STT~\cite{zhang2021spatiotemporal} & 86.3             & 88.7             & --               & --            \\ \hline
			Our                & \textbf{LSTRL}                             & \textbf{86.8} & \textbf{91.6} & \underline{92.2}    & \textbf{98.6}    \\ \hline
		\end{tabular}%
	}
	\vspace{-4mm}
\end{table}

\section{Experiments}
\subsection{Datasets and Evaluations}
In this paper,  we adopt three widely-used V-ReID benchmarks, \emph{i.e.}, iLIDS-VID~\cite{wang2014person}, MARS~\cite{zheng2016mars} and LS-VID~\cite{li2019global}, to evaluate our proposed method.
iLIDS-VID is a small-scale dataset with 600 video sequences of 300 different identities.
MARS is one of the large-scale datasets and consists of 1,261 identities around 18,000 video sequences.
LS-VID is another large-scale dataset.
It comprises 4,832 sequences from 1,812 identities.
Follow previous works~\cite{gu2020appearance,wang2021pyramid}, we compute the cumulative matching characteristic table, including R-1 and R-5, and mean Average Precision (mAP) for algorithm evaluation.
\subsection{Implement Details}
We conduct experiments with the Pytorch toolbox.
Our experimental devices include an Intel i4790 CPU and two NVIDIA RTX3090 GPUs (24G memory).
During training, we adopt the Restricted Random Sampling (RRS)~\cite{li2018diversity} to sample 8 frames from one video as a sequence input.
Each frame is resized to 256$\times$128 and augmented by random cropping and erasing.
In a mini-batch, 8 person identities are selected randomly, and each identity samples 4 video sequences.
The Adam~\cite{kingma2014adam} algorithm is adopted to optimize the whole framework.
Its initial learning rate is 0.0003 and decayed by 10 at every 70 epochs until 400 epochs.

\subsection{Comparison with State of the Arts}
We compare our LSTRL with state-of-the-art methods. 
The results are presented in Table~\ref{tab:Sota1} and~\ref{tab:Sota2}.
It is observed that our method outperforms the most state-of-the-art methods on MARS and iLIDS-VID datasets, and attains the highest retrieval accuracy on LS-VID dataset.

In the cases of spatial appearance learning, some previous methods concentrate on a single frame and extract the salient features, such as DRSA~\cite{li2018diversity}, BiCNet~\cite{hou2021bicnet} and SINet~\cite{bai2022salient}.
In particular, SINet attains $91.0\%$ ,$92.5\%$ and $87.4\%$ R-1 accuracy on MARS, iLIDS-VID and LS-VID, respectively.
Better than SINet, our method extracts the long-term appearance representations from all the frames and get higher retrieval accuracies on MARS and LS-VID.
Noted that, COSAM~\cite{subramaniam2019co} aggregates the frame-level features to activate spatial meaningful regions, and RAFA~\cite{zhang2020multi} learns multi-granularity representations from single-frame features.
Both of them ignore the inter-frame associations.
Different from them, we capture four kinds of dependencies to associate inter-frame features, which helps to extract long-term appearance representations.
Thus, our method outperforms COSAM and RAFA by $6.9\%$ and $2.8\%$ in terms of mAP and R-1 on MARS.
\begin{table}[t!]
	\centering
	\caption{Performance (\%) comparison with state-of-the-arts on LS-VID dataset.}
	\label{tab:Sota2}
	\resizebox{0.38\textwidth}{!}{%
		\begin{tabular}{c|l|cc}
			\hline
			\multicolumn{1}{l|}{} & \multirow{2}{*}{Methods}           & \multicolumn{2}{c}{LS-VID}          \\
			&                                    & mAP              & R-1              \\ \hline
			\multirow{2}{*}{S}    & BiCNet~\cite{hou2021bicnet}        & 75.1             & 84.6             \\
			& SINet~\cite{bai2022salient}        & \underline{79.6} & \underline{87.4 }            \\ \hline
			\multirow{3}{*}{T}    & GLTR~\cite{li2019global}           & 44.3             & 63.1             \\
			& TCLNet~\cite{hou2020temporal}      & 70.3             & 81.5             \\ \hline
			\multirow{2}{*}{ST}   & STMN~\cite{eom2021video}           & 69.2             & 82.1             \\
			& STT~\cite{zhang2021spatiotemporal} & 78.0             & 87.5             \\ \hline
			Our                   & \textbf{LSTRL}                     & \textbf{82.4}    & \textbf{89.8}    \\ \hline
		\end{tabular}%
	}
\end{table}

Meanwhile, some works, such as GRL~\cite{liu2021watching}, TCLNet~\cite{hou2020temporal} and GLTR~\cite{li2019global}, deploy temporal learning to associate and aggregate frame-level features.
However, these works hardly capture the motion information of persons.
Thus, for the extraction of motion representations, previous methods, such as RNN$+$Flow~\cite{mclaughlin2016recurrent} and AMOC$+$Flow~\cite{liu2017video}, calculate the optical flow and combine RGB images as inputs.
%
%
Although effective, it brings large computational complexity. 
Besides, some recent works, such as AMEM~\cite{li2020appearance}, model the walking patterns of pedestrians to capture the motion features.
%
In our work, we perform the global-to-local and reciprocal motion extraction between consecutive frames, which consumes only a small computational cost.
Combined short-term motion and long-term appearance representations, our method can surpasses AMEM by $7.5\%$ mAP and $4.9\%$ R-1 on MARS, respectively.

Moreover, AP3D~\cite{gu2020appearance}, CTL~\cite{liu2021spatial} and STT~\cite{zhang2021spatiotemporal} model spatial-temporal information.
Different from them, our method effectively extracts long-term appearance and short-term motion representations.
In addition, the proposed MAE and BME can be easily inserted into existing networks for efficient learning.
Results show that our method attains better performances than AP3D, CTL and STT.

\begin{table}[t!]
	\parbox{1.0\linewidth}{
		\centering
		\caption{Ablation results (\%) of key modules on MARS and LS-VID datasets.}
		\label{tab:KeyModule}
		\resizebox{0.55\textwidth}{!}{%
			\begin{tabular}{c|cc|cc|c|c}
				\hline
				\multirow{2}{*}{Method} &
				\multicolumn{2}{c|}{MARS} &
				\multicolumn{2}{c|}{LS-VID} &
				\multirow{2}{*}{\begin{tabular}[c]{@{}c@{}}Param.\\ (M)\end{tabular}} &
				\multirow{2}{*}{\begin{tabular}[c]{@{}c@{}}FLOPs\\ (G)\end{tabular}} \\
				& mAP  & R-1  & mAP  & R-1  &  &  \\ \hline
				Baseline & 84.2 & 89.4 & 77.8 & 86.9 & 24.79 & 4.09 \\
				+ MAE    & 86.3 & 90.9 & 81.2 & 89.0  & 26.43 & 4.42 \\
				+ BME    & 86.8 & 91.6 & 82.4  & 89.8  & 29.06 & 4.96 \\ \hline
			\end{tabular}%
		}
	}

	\hfill
	\parbox{.46\linewidth}{
		\centering
		\vspace{5mm}
		\caption{Ablation results (\%) of multi-granularity appearances on MARS and LS-VID datasets.}
		\label{tab:MAE}
		\resizebox{0.43\textwidth}{!}{%
			\begin{tabular}{c|cc|ll}
				\hline
				\multirow{2}{*}{Mehods} & \multicolumn{2}{c|}{MARS} & \multicolumn{2}{c}{LS-VID}                        \\ \cline{2-5}
				& mAP         & R-1         & \multicolumn{1}{c}{mAP} & \multicolumn{1}{c}{R-1} \\ \hline
				Baseline+MAE            & 86.3        & 90.9        & 81.2                    & 89.0                    \\ \hline
				- $\textbf{A}_1$        & 85.9        & 90.6        & 80.8                    & 88.4                    \\
				- $\textbf{A}_2$        & 85.7        & 89.9        & 80.5                    & 88.2                    \\
				- $\textbf{A}_3$        & 85.9        & 90.2        & 80.0                    & 88.3                    \\
				- $\textbf{A}_4$        & 85.7        & 90.1        & 80.3                    & 88.6                    \\ \hline
			\end{tabular}%
		}
	}
	\hfill
	\parbox{.5\linewidth}{
		\centering
		\caption{Ablation results (\%) of Motion Estimation on MARS and LS-VID Datasets.}
		\vspace{2mm}
		\label{tab:BME}
		\resizebox{0.5\textwidth}{!}{%
			\begin{tabular}{cc|cc|cc|cc}
				\hline
				\multicolumn{2}{c|}{Manner} & \multicolumn{2}{c|}{Direction} & \multicolumn{2}{c|}{MARS} & \multicolumn{2}{c}{LS-VID} \\
				Local-       & Global-      & Single-      & Bi-          & mAP  & R-1  & mAP  & R-1  \\ \hline
				$\checkmark$ &              &              & $\checkmark$ & 86.1 & 90.7 & 82.0 & 88.8 \\
				& $\checkmark$ & $\checkmark$ &              & 86.7 & 91.0 & 82.2 & 89.4 \\
				& $\checkmark$ &              & $\checkmark$ & 86.8 & 91.6 & 82.4 & 89.8 \\ \hline
			\end{tabular}%
		}
	}
\end{table}

\subsection{Ablation Study}
In this section, we conduct more experiments to investigate the effectiveness of our proposed modules.

\textbf{Effectiveness of MAE and BME.}
We carry out incremental validation on MARS and LS-VID datasets.
The experimental results are shown in Table~\ref{tab:KeyModule}.
Baseline refers to the model only using ResNet-50.
Then, we insert MAE into ResNet-50 at the end of the Res2 and Res3 blocks respectively.
When deploying MAE, the frame-level features get enhanced.
Compared with the baseline method, our MAE significantly improves the mAP by $5.7\%$ and $2.8\%$ on MARS and LS-VID, respectively.
The improvements indicate that the extraction of long-term appearances is effective.
In addition, two BMEs are deployed at the same layers and promote to gain further growth of $0.7\%$ and $0.8\%$ R-1 on MARS and LS-VID.
These results validate the complementarity of short-term motions and long-term appearances.
Meanwhile, the model complexity (Param.) and computational complexity (FLOPs) are reported in Table~\ref{tab:KeyModule}.
One can see that the computational cost increase is relatively small, showing that our long short-term representation learning is efficient.

\textbf{Necessities of Multi-granularity Appearance.}
Table~\ref{tab:MAE} shows the performances of MAE on MARS and LS-VID when removing one of the four-granularity appearances.
We find that the accuracies have decreased in different degrees.
Especially, when the appearance feature $\textbf{A}_2$ is missing, the accuracies of R-1 and mAP drop $0.6\%$  and $1.0\%$ on MARS respectively.
These findings account for the necessity of long-term appearances in multi-granularity.
\begin{figure}[t]
	\centering
	\resizebox{0.95\textwidth}{!}
	{
		\begin{tabular}{@{}c@{}c@{}}
			\includegraphics[width=\linewidth,height=0.7\linewidth]{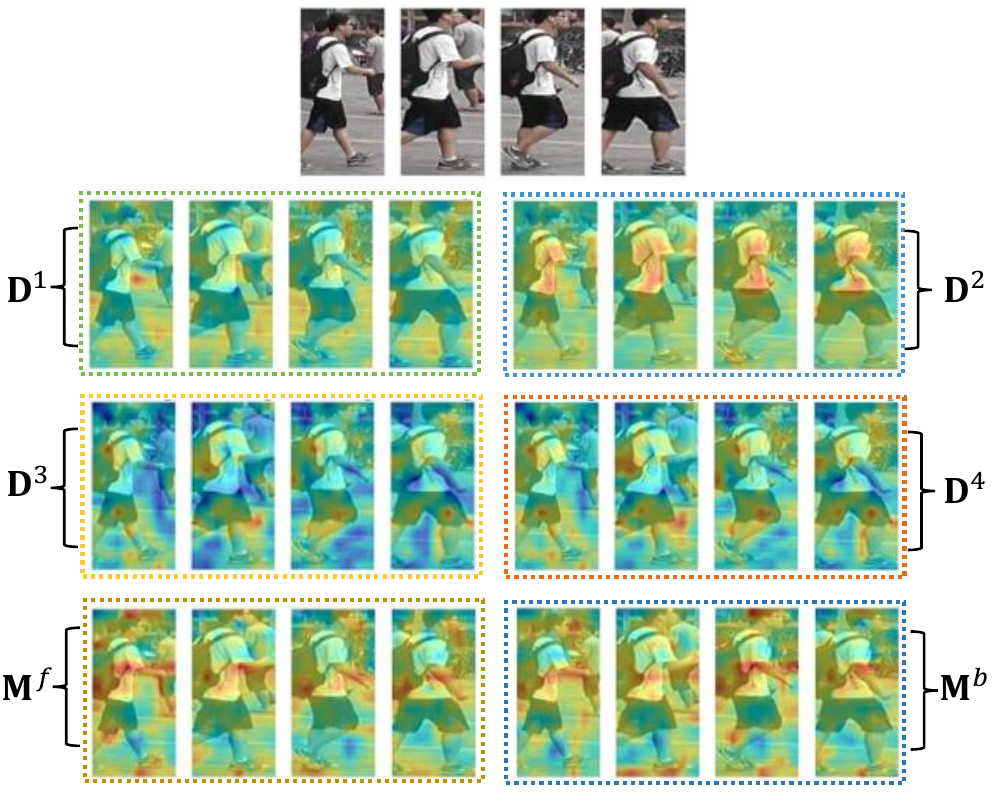} \\
		\end{tabular}
	}
	\vspace{-4mm}
	\caption{The visualizations of multi-granularity dependencies ($\textbf{D}^1$, $\textbf{D}^2$, $\textbf{D}^3$, $\textbf{D}^4$) in MAE, and the motion features in forward ($\textbf{M}^f$) and backward ($\textbf{M}^b$) directions from BME.
	}
	\label{fig:Visualization}
	\vspace{-4mm}
\end{figure}

\textbf{Effects of Motion Estimation.}
In our BME, we perform global-to-local and bi-direction estimation to obtain short-term motion representations.
As shown in Table~\ref{tab:BME}, compared with the local-to-local estimation, our global-to-local estimation gains better retrieval accuracies.
In the deployment of local-to-local motion estimation, the dot product is conducted between any pair-wise local features.
The reciprocal temporal learning actually increases the robustness to noise frames.
Thus, as shown in Table~\ref{tab:BME}, our bi-direction motion estimation further raises the R-1 by $0.6\%$  and $0.4\%$ on MARS and LS-VID than single-direction.

\textbf{Visualization Analysis.}
In Fig.~\ref{fig:Visualization}, we visualize the multi-granularity dependencies, which are captured by MAE and present the inter-frame associations.
One can see that $\textbf{D}^1$ associates each local feature.
$\textbf{D}^2$ highlights the salient regions in spatial, such as the upper body.
$\textbf{D}^3$ and $\textbf{D}^4$ tend to mine more meaningful cues under global guidance.
Meanwhile, the motion feature maps $\textbf{M}^f$ and $\textbf{M}^b$ obtained by BME are also visualized. 
Comparing $\textbf{M}^f$ and $\textbf{M}^b$, it can be seen that, both of them focus on the main body of persons, but there are some differences in the highlighted areas, which validate the benefits of the bi-direction estimation.
Overall, these visualization results further demonstrate the effectiveness of our MAE and BME.
\section{Conclusion}
In this paper, we propose a novel learning framework for video-based person Re-ID.
To extract long-term representations, we propose the MAE, in which four granularity appearances are effectively captured by associating multiple frames.
To extract short-term representations, we propose the BME, in which reciprocal motion information is efficiently estimated from consecutive frames.
Our MAE and BME are plug-and-play and can be easily inserted into existing networks for end-to-end supervision.
As a result, they significantly improve the feature representation ability for video-based person Re-ID.
Experiments on three public benchmarks show that our approach outperforms most state-of-the-arts.

\bibliographystyle{splncs04}
\bibliography{ICIG_LSTRL}

\end{document}